\pdfoutput=1
\documentclass[11pt]{article}

\usepackage[]{acl}

\usepackage{times}
\usepackage{latexsym}
\usepackage{multirow}
\usepackage{graphicx}
\usepackage{tabularx}
\usepackage{booktabs}
\usepackage{amsmath}

\usepackage[T1]{fontenc}
\usepackage[utf8]{inputenc}

\usepackage{microtype}

\title{Distribution Calibration for Out-of-Domain Detection with Bayesian Approximation}

\author{Yanan Wu$^{1*}$, Zhiyuan Zeng$^{1*}$, Keqing He$^{2*}$, Yutao Mou$^{1}$ , Pei Wang$^{1}$, {\bf Weiran Xu$^{1}$}\thanks{\ \ The first three authors contribute equally. Weiran Xu is the corresponding author.}\\
  $^1$Beijing University of Posts and Telecommunications, Beijing, China\\
$^{2}$Meituan Group, Beijing, China\\
  \texttt{\{yanan.wu,zengzhiyuan,myt,wangpei,xuweiran\}@bupt.edu.cn}\\
  \texttt{\{hekeqing\}@meituan.com}
  }
  
\begin{document}
\maketitle
\begin{abstract}
% Out-of-Domain (OOD) detection is a key component in a task-oriented dialog system. It aims to identify whether a query falls outside the predefined supported intent set. Previous softmax-based detection algorithms are proved to be overconfident for OOD samples. In this paper, we first perform a theoretical analysis of the reason behind the overconfidence issue and find a severe mismatch between hypothesis OOD distribution and practical distribution. Then we propose a simple but strong Bayesian approximation method to calibrate the practical sparse OOD distribution over the simplex into ideal dense OOD distribution for OOD detection. We aim to make the expectation of each OOD softmax prediction distribution similar to a uniform distribution. Experiments and analysis show our method gains significant improvements over softmax-based baselines.

Out-of-Domain (OOD) detection is a key component in a task-oriented dialog system, which aims to identify whether a query falls outside the predefined supported intent set. Previous softmax-based detection algorithms are proved to be overconfident for OOD samples. In this paper, we analyze overconfident OOD comes from distribution uncertainty due to the mismatch between the training and test distributions, which makes the model can't confidently make predictions thus probably causing abnormal softmax scores. We propose a Bayesian OOD detection framework to calibrate distribution uncertainty using Monte-Carlo Dropout. Our method is flexible and easily pluggable into existing softmax-based baselines and gains 33.33\% OOD F1 improvements with increasing only 0.41\% inference time compared to MSP. Further analyses show the effectiveness of Bayesian learning for OOD detection. \footnote{Our code is available at \url{https://github.com/pris-nlp/COLING2022_Bayesian-for-OOD/}.}
\end{abstract}

\vspace{-0.1cm}
\section{Introduction}
\vspace{-0.1cm}
% 1. 介绍OOD的定义
% 2. 介绍OOD的常用方法 MSP 、Entropy，重点突出他们面临的overconfidence问题，引出后续的理论假设
% 3. 从理论的角度阐述overconfidence现象背后的原因，结合fig说明
% 4. 介绍我们的方法，重点是为什么我们的方法可以work，背后的理论解释
% 5. 贡献

% Update:
% 1. 第三段理论分析要移到方法，并且尽量避免 理论 这个词，换成一个具体的示意图（贝叶斯校准前后概率分布的变化）

% 输入sentence：adjust the contrast of the black and white photos【GT=OOD】
% (seed=1,2,3,4的结果 以及 取平均后的结果)
% 预测标签（max softmax score）: "change_volume"(0.76),"share_location"(0.73),"smart_home"(0.57),"change_accent"(0.56),"unseen"(0.19)

% 输入sentence：block my american saving bank for now【GT=IND】(其实也存在IND预测错误，平均后纠正的情况)
% (seed=1,2,3,4的结果 以及 取平均后的结果)
% 预测标签（max softmax score）: 'freeze_account'(0.79)(0.84)(0.91)(0.95)(0.90)

Detecting Out-of-Domain (OOD) or unknown intents from user queries is key for a task-oriented dialog system \cite{Gnewuch2017TowardsDC,Akasaki2017ChatDI,Tulshan2018SurveyOV,Shum2018FromET,zeng-etal-2021-modeling,zeng-etal-2021-adversarial,wu-etal-2022-revisit}. It aims to know when a user query falls outside their range of predefined supported intents to avoid performing wrong operations. Different from normal intent classification tasks, lack of labeled OOD examples leads to poor prior knowledge about these unknown intents, making it challenging to detect OOD samples.

A rich line of OOD intent detection algorithms has been developed recently, among which softmax-based methods demonstrated promise \cite{Guo2017OnCO,Liang2018EnhancingTR,Zheng2020OutofDomainDF}. Softmax-based methods leverage softmax outputs extracted from an in-domain (IND) intent model and operate under the assumption that the test OOD samples get a lower likelihood probability than the ID data. For example, Maximum Softmax Probability (MSP) \cite{Hendrycks2017ABF} detects a test query as OOD if its max softmax probability is lower than a fixed threshold. However, all these models make a strong distributional assumption of the practical OOD probability being uniform, which has been proven wrong because neural networks can produce over-confidently high softmax scores even for OOD samples \cite{Guo2017OnCO}. Therefore, solving the overconfidence issue is still challenging for OOD detection.

% \begin{figure}[H]
% \begin{minipage}[t]{1\linewidth}
%   \centering
%   \includegraphics[width=3in,height=2in]{x5.eps}\\
% \centering{(a) 子图标题}\\
%  \end{minipage}%

\begin{figure}[t]
    \centering
    \resizebox{.48\textwidth}{!}{
    \includegraphics{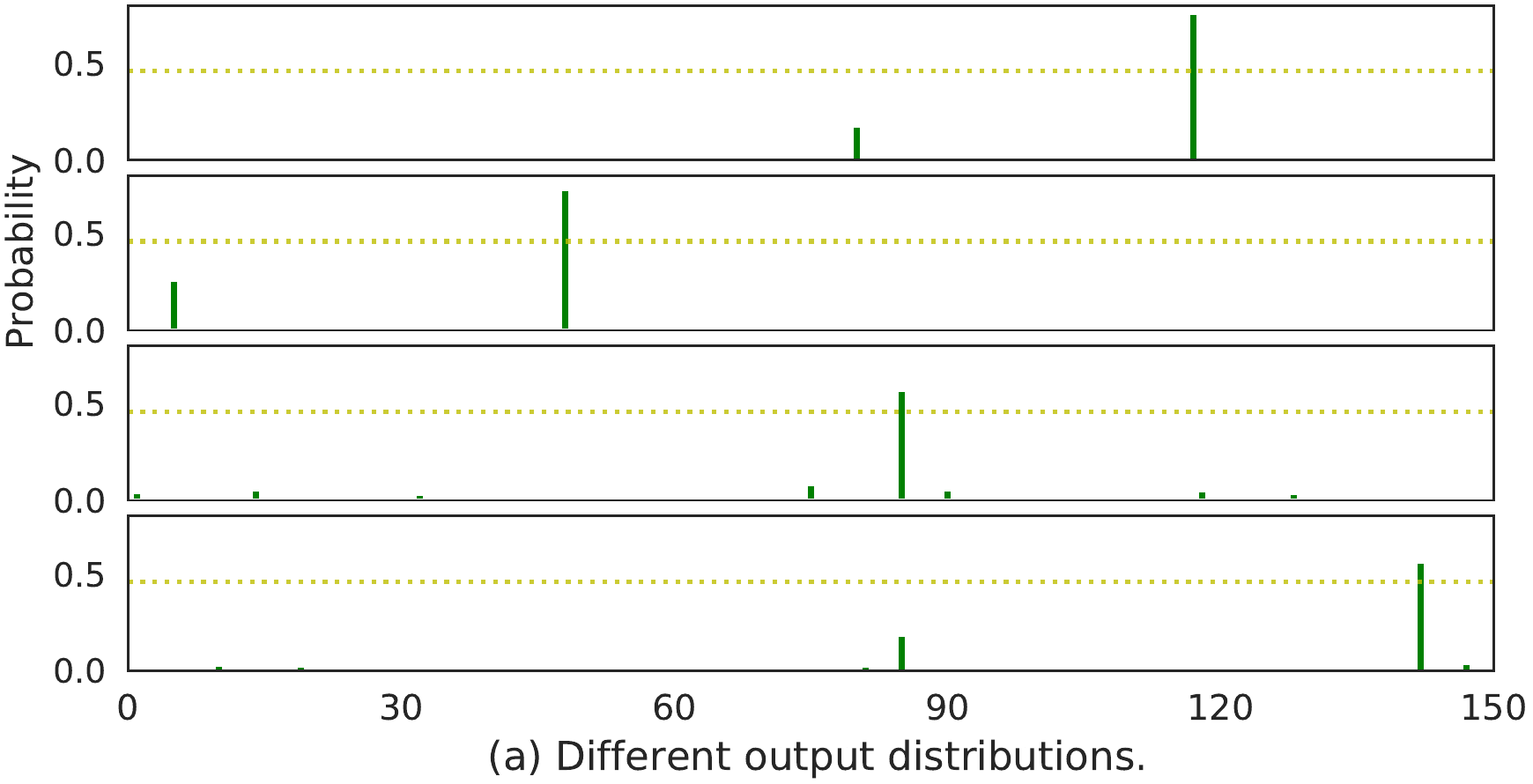}}
    \vspace{-0.3cm}
    \label{intro_unseen_before}
     \vspace{-0.7cm}
\end{figure}
\begin{figure}[t]
    \centering
    \resizebox{.48\textwidth}{!}{
    \includegraphics{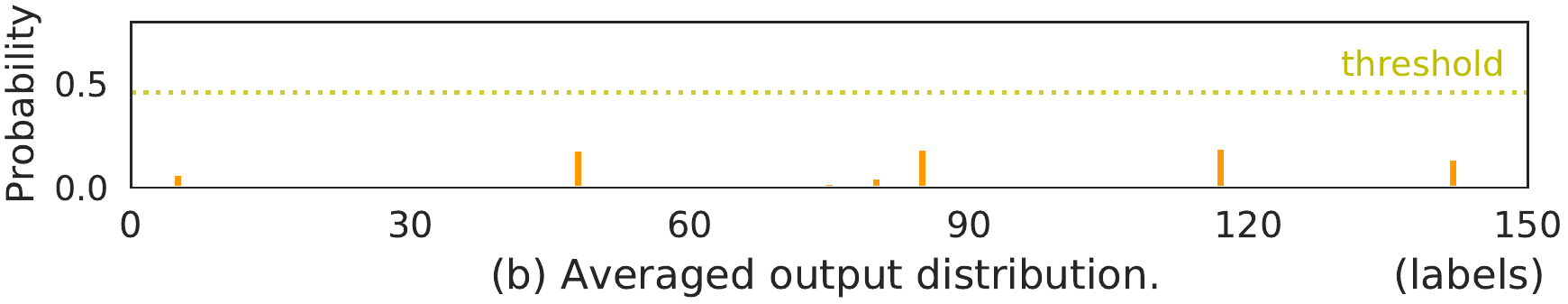}}
    \vspace{-0.7cm}
    \caption{Multiple predicted probability distributions of an OOD sample under different random seeds. We train four identical models on the same data but only use different random seeds. Fig (a) displays each output distribution of an OOD input and Fig (b) shows the averaged output distribution.}
    \label{intro_unseen_after}
     \vspace{-0.75cm}
\end{figure}

In this paper, we study the overconfidence issue from the perspective of Bayesian learning \cite{pmlr-v48-gal16}. Essentially, the reason for overconfidence is that a model cannot confidently make predictions on the input OOD utterances (unknow-unknow) due to the lack of prior knowledge of OOD data. In other words, even given the same input OOD intent, the predicted probability distributions using different random seeds are completely different, uniform, sharp, or any distribution. Fig \ref{intro_unseen_after} show an example. We find models with different initialization seeds can output diverse distributions for OOD input, maybe cause overconfidence in several in-domain classes. But the averaged output is close to a uniform distribution. We also find models with different seeds are more robust to IND input and obtain consistent outputs (see Appendix \ref{ind_random}). Therefore, one direct way to solve the distribution uncertainty is to train multiple models independently and assemble their outputs for the final result. But this method is not applicable to practical scenarios for large training cost. In this paper, we propose a Bayesian OOD detection framework to calibrate distribution uncertainty. Specifically, we firstly train an in-domain intent classifier using IND data, then in the test stage, we perform multiple stochastic forward passes with a certain dropout rate (like 0.7) and average the output normalized logits as a final probability. Without increasing any new parameters, we calibrate distribution uncertainty by tending to expectation uniform distribution via Monte-Carlo Dropout \cite{pmlr-v48-gal16}. Our method can be easily extended to existing softmax-based OOD detection methods and gain significant OOD improvements with only increasing little inference time compared to baselines, even outperform the state-of-the-art distance-based methods like LOF \cite{Lin2019DeepUI} and GDA \cite{xu-etal-2020-deep}. Our contributions are two-fold: (1) We analyze the intrinsic reason of overconfidence issue via distribution uncertainty and propose a Bayesian OOD detection framework to calibrate this uncertainty using Monte-Carlo Dropout. (2) We provide theoretical and empirical analysis to demonstrate the effectiveness of our Bayesian OOD method.

% In this paper, we propose a simple but strong OOD detection algorithm to mitigate the overconfidence issue of softmax-based methods via Bayesian approximation. We simply adopt Monte-Carlo Dropout \cite{pmlr-v48-gal16} and average the output logits from multiple stochastic forward passes. Then we use the average output as the final softmax confidence score like MSP. The intuition is that expectation of each OOD softmax prediction distribution is equal to a uniform distribution and our Bayesian approximation can calibrate the practical sparse OOD distribution (green dots) over the simplex into ideal dense OOD distribution (yellow dots) to match the theory hypothesis of the softmax-based detection algorithms like MSP. 
% Our contributions are three-fold: (1) We analyze the reason behind the overconfidence issue of softmax-based detection algorithms and obtain the theoretical explanation. (2) Based on the theoretical analysis, we propose a simple but strong Bayesian approximation method to mitigate the overconfidence issue for OOD detection. (3) Extensive experiments and comprehensive analysis on two public datasets demonstrate the effectiveness of our method.

\begin{figure*}[t]
    \centering
    \resizebox{.9\textwidth}{!}{
    \includegraphics{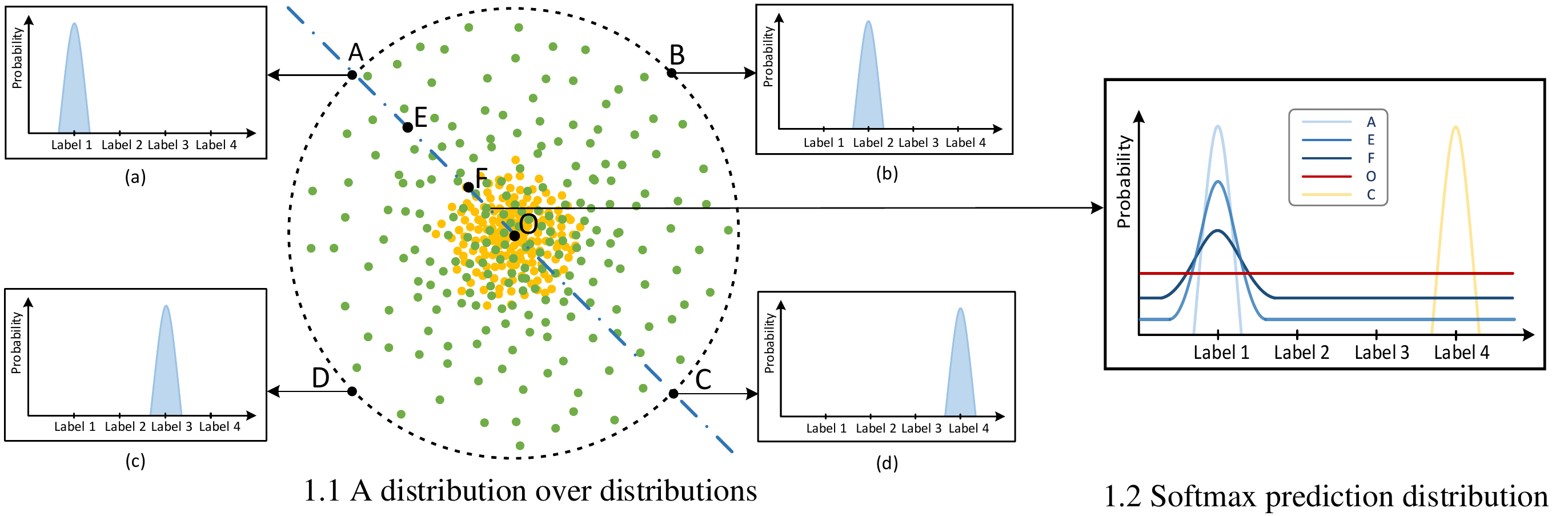}}
    \vspace{-0.3cm}
    \caption{A distribution over distributions where each dot represents a softmax prediction distribution for a test OOD sample and all the dots denote a distribution over distributions. We also display softmax prediction distributions of several dots on the dot line in Fig 1.2.}
    \label{intro}
     \vspace{-0.4cm}
\end{figure*}

\vspace{-0.2cm}
\section{Method}
\vspace{-0.1cm}
\subsection{Understanding OOD Detection}
\vspace{-0.1cm}
% 引用不确定性理论：建模两部分，拆解为三部分
% 说明三部分不确定性都代表什么？
%   - 模型不确定性，数量大可缓解
%   - 数据不确定性，know-unknow，即黄点，与理想假设一致；
%   - 分布不确定性，unknow-unknow。即在不同随机数下，对同一个OOD样本的预测概率分布差异大（对应intro图），导致在图2中OOD呈现绿色弥散点现象，与理想分布不一致；也就是对于OOD样本来说，呈现overcondidence的现象。
%  因此，解决overconfidence问题关键是如何缓解分布的不确定性，使得OOD样本点从真实的绿色点转换为黄色理想稠密的点，缓解过自信问题。

% 考虑边缘化\theta；利用MC dropout近似（dropout的数学表达式）；得到最终我们方法的公式；

% 假设1：理想OOD样本的概率分布呈现均匀分布，所以有MSP算法；理想IND样本的概率分布是sharp分布；
    % 证明1: 先前工作
    % 证明2: OOD不属于任何一个类别
    % 证明3: 实验：样本-level的概率分布，数据集-level的KL散度
% 那么根据大数定理，当我们采样次数够大时，则会趋近于必然事件，也就是OOD样本的概率分布呈现均匀分布，IND样本的概率分布呈现sharp分布；也就是说我们可以通过贝叶斯估计校准概率分布。

% 定义符号 D, x, y, OOD任务
\textbf{Problem Definition} We refer to training data $D$ as IND data. We aim to detect the input utterances $x$ belonging to OOD and correctly classify the utterances belonging to IND utilizing a well-calibrated classifier trained only on finite IND data $D$. %We aims to detect the input utterance $x$ is from $V_{ind}$ or $V_{ood}$ utilizing a well calibrated classifier $P(y|x,D)$ trained only on finite IND data $D$. and each utterance in D is drawn from a fixed but unknow distribution $V_{ind}$. We let $V_{ood}$ denote an out-of-distribution (OOD) which is 'far away' from in-distribution $V_{ind}$.

% 讨论三个不确定性，引出解决OOD的过自信问题的关键在于分布不确定性。
The predictive uncertainty of a classification model $P(\upsilon|x,D)$ is commonly divided into  \emph{data uncertainty} (aleatoric), \emph{distribution uncertainty} and \emph{model uncertainty} (epistemic)\cite{twouncertainty2009, Malinin2018PredictiveUE}: 

\begin{equation}
\setlength{\abovedisplayskip}{0.05cm}
\setlength{\belowdisplayskip}{0.01cm}
P(\upsilon|x,D) = \iint{\underbrace{P(\upsilon|\mu)}_{\text{data}}\underbrace{P(\mu|x,\theta)}_{\text{distribution}}\underbrace{P(\theta|D)}_{\text{model}}d\mu\!\,d\theta}
\label{eq2}
\end{equation}

The \textbf{model uncertainty} is described by the posterior distribution over model parameters $\theta$, and it can be lowered by increasing the amount of data and simplifying the model complexity. 
The \textbf{data uncertainty} is described by the posterior distribution over classes, where $\upsilon$ is the predicted distribution of all possible in-domain intent classes for OOD detection. It arises from the natural complexity of the data, such as class overlap, label noise and homoscedastic noise.  It is a property of the world, and cannot be changed. % Inspired by Dirichlet \cite{Malinin2018PredictiveUE}, we consider the distribution of the distribution of IND and OOD, and split Eq \ref{eq1} into the following: % \mu 指的是单纯性的分布
The \textbf{distribution uncertainty} is modeled with a distribution over distribution, where $\mu$ is the categorical distribution over simplex. It arises due to the mismatch between the training and test distributions. 
We give an example in Fig \ref{intro} which displays a distribution over distributions on a simplex (Dirichlet distribution \cite{Malinin2018PredictiveUE}) where each dot represents a softmax prediction distribution for a test OOD sample and all the dots denote a distribution over distributions. For an input utterance $x$, softmax-based detection algorithms like MSP assume that the distribution of OOD utterances ($v_{ood}$) should be very close to the uniform distribution (the yellow dots in Fig \ref{intro}) and the distribution of IND utterances ($v_{ind}$) should be very close to one-hot distribution (e.g. Fig \ref{intro}(a)-(d)). However, the practical OOD samples (green dots) exactly yield a sparse distribution over the simplex where each OOD sample may get a sharp softmax prediction distribution (like one-hot distribution) or a flat softmax prediction distribution (like uniform distribution). Due to the lack of prior knowledge of OOD data, the model cannot confidently make predictions on the input OOD utterances (unknow-unknow) which is the essential reason why the predicted probability distributions of the same OOD sample are completely different under different random seeds and even get very high max softmax scores (Fig \ref{intro_unseen_after}). In other words, distribution uncertainty could lead to overconfidence in the prediction of OOD samples. 

Therefore, how to alleviate the distribution uncertainty is the key to solving the overconfidence problem in OOD detection. %In this paper, we aim to calibrate the practical sparse OOD distribution (green dots) over the simplex into ideal dense OOD distribution (yellowdots) to mitigate the overconfidence issue.

% 贝叶斯估计
% \subsection{Distribution Calibration}
\vspace{-0.1cm}
\subsection{Bayesian Approximation}
\vspace{-0.1cm}
In order to alleviate the distribution uncertainty, we consider marginalizing out $\theta$ in Eq \ref{eq2}:
\vspace{-0.5cm}
\begin{equation}
P(\upsilon|x,D) = \int{P(\upsilon|\mu)P(\mu|x,D)d\mu}
\label{eq3}
\vspace{-0.09cm}
\end{equation}
This yields expected estimates of data and distributional uncertainty given model uncertainty. Marginalization is intractable in deep neural networks, thus we consider using $q(\omega)$ to approximate the intractable posterior though Monte-Carlo Sampling algorithm \cite{Tsymbalov2020DropoutSB,pmlr-v48-gal16}:
\begin{equation}
P(\upsilon|x,D) = \int{P(\upsilon|\omega)q(\omega)d\omega}
\label{eq3}
\end{equation}
where $\omega = \{W_i\}_{i=1}^l$ is the random variables for a model with $l$ layers. We define $q(\omega)$ as:
\begin{equation}
W_{i} = M_i \cdot diag(\,\![\alpha_{i,j}\!\,]_{i=1}^{k_i})
\label{eq4}
\end{equation}
\begin{equation}
\alpha_{i,j} \sim Bernoulli(p_i)
\label{eq5}
\end{equation}

Where $p_i$ and $M_i$ are the variational parameters. The binary variable $\alpha_{i,j}$ indicates whether unit $j$ of the $i-1$ layer will be passed to the next layer. % Fig \ref{model} shows the Bayesian approximation process for distribution calibration. 
Specifically, we sample $N$ sets of independent random vectors of realisations from the Bernoulli distribution $\{\alpha_1^n,...,\alpha_l^n\}_{n=1}^N$ with $\![\alpha_{i,j}\!]_{i=1}^{k_i}$ giving $\{W_1^n,...,W_L^n\}_{n=1}^{N}$. Then we average the output:% normalized
\begin{equation}
E_{q(\upsilon|x,D)}\!(\upsilon\!)\approx \frac{1}{N} \sum_{n=1}^N \hat \upsilon{(x,D,W_1^n,...,W_l^n)}
\label{eq5}
\end{equation}
According to the Law of Large Numbers \cite{Yao2016LawOL}, when N is large enough, the predicted distribution will converge in expected uniform distribution. That is, we can calibrate the practical sparse OOD distribution (green dots) over the simplex into ideal dense OOD distribution (yellow dots) by Bayesian approximation to mitigate the overconfidence issue, which is verified in the following empirical experiments.
% the predicted distribution will converge in expected uniform distribution
% when N is large enough, the sample mean will converge in probability to the true mean and the probability of occurrence of random events can be corrected to the expected probability.

\subsection{OOD Detection with Bayesian Learning}
\begin{figure}[t]
    \centering
    \resizebox{.46\textwidth}{!}{
    \includegraphics{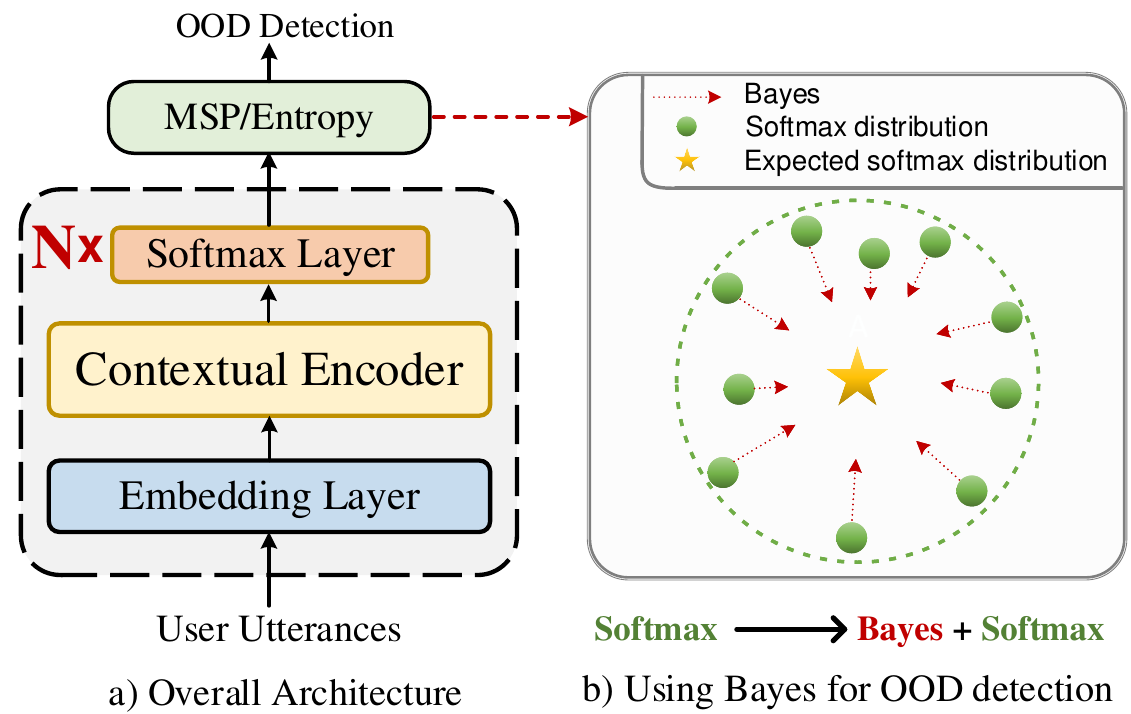}}
    \vspace{-0.3cm}
    \caption{The overall architecture of our method.}
    \label{model}
     \vspace{-0.4cm}
\end{figure}

Fig \ref{model}(a) shows the overall architecture of our proposed OOD detection model. The part in the dashed box is a well-trained feature extractor based on Bi-LSTM \cite{Hochreiter1997LongSM} or BERT \cite{Devlin2019BERTPO}. It is trained on labeled in-domain data using cross-entropy loss. Fig \ref{model}(b) shows the Bayesian approximation process for distribution calibration. We adopt Monte-Carlo Dropout and average the output normalized logits from multiple stochastic forward passes: $\boldsymbol{\bar{v}} = \frac{1}{N}\sum_{i=1}^N\boldsymbol{v_i}$. In this way, we calibrate the softmax distribution to the expected distribution, which close to a uniform distribution. Then, we apply two softmax-based metrics for OOD detection, which is $m_{MSP} = \max(\boldsymbol{\bar{v}})$ and $m_{Entropy} = -\sum_{i=1}^c \bar{v}_i\log \bar{v}_i$. We further apply a empirical threshold to distinguish IND and OOD data.

\begin{table*}[t]
\normalsize
\centering
\resizebox{0.85\linewidth}{!}{
\begin{tabular}{l|l|c|c|c|c|l|l|l|l}
\toprule[1pt]
\multicolumn{2}{l|}{\multirow{3}{*}{Model}} & \multicolumn{4}{c|}{CLINC-Full}                                   & \multicolumn{4}{c}{CLINC-Imbal}                                                                          \\ \cline{3-10} 
\multicolumn{2}{l|}{}                       & \multicolumn{2}{c|}{OOD}        & \multicolumn{2}{c|}{IND}        & \multicolumn{2}{c|}{OOD}                              & \multicolumn{2}{c}{IND}                          \\ \cline{3-10} 
\multicolumn{2}{l|}{}                       & F1             & Recall         & F1             & ACC            & \multicolumn{1}{c|}{F1} & \multicolumn{1}{c|}{Recall} & \multicolumn{1}{c|}{F1} & \multicolumn{1}{c}{ACC} \\ \hline
\multirow{6}{*}{LSTM}   & LOF \cite{Lin2019DeepUI}              & 59.28          & 58.32          & 86.08          & 85.87          &55.37  &51.03  &80.51  &82.79\\ 
                        & GDA \cite{xu-etal-2020-deep}              & 65.79          & 64.14          & 87.90          & 86.83          &61.38 	&63.80 	&85.35 	&84.20         \\ \cline{2-10} 
                        & MSP  \cite{Hendrycks2017ABF}             & 50.13          & 45.60          & 87.73          & 87.25          &44.93  &41.10  &84.96  &84.16                   \\ 
                        & MSP+Bayes.(ours)  & 70.05          & 68.38          & 88.91          & 88.57          &61.70  &57.50  &85.92  &85.65                         \\ \cline{2-10} 
                        & Entropy  \cite{Zheng2020OutofDomainDF}        & 68.05          & 67.96          & 88.97          & 88.68          &64.45  &63.80  &86.07  &85.71                   \\ 
                        & Entropy+Bayes.(ours)   & \textbf{72.02} & \textbf{71.70} & 89.10 & 88.73 &\textbf{68.32}   &\textbf{67.61}   &86.34   &86.11                      \\ \hline\hline
\multirow{4}{*}{BERT}   & MSP               & 52.79          & 50.50           & 87.81          & 87.46          &48.76  &46.70  &85.87  &85.65                        \\ 
                        & MSP+Bayes.(ours)  & 71.25          & 69.58          & 89.10          & 89.56          &64.32  &62.00  &86.39  &85.87                       \\ \cline{2-10} 
                        & Entropy           & 68.97          & 68.83          & 89.13          & 88.72         &65.25  &64.89  &86.21  &85.94                     \\  
                        & Entropy+Bayes.(ours)   & \textbf{72.85} & \textbf{72.42} & 89.47 & 88.94 &\textbf{69.11}   &\textbf{68.49}   &86.74   &86.42                        \\ \bottomrule[1pt]
\end{tabular}}
\vspace{-0.2cm}
\caption{Performance comparison between our method and baselines on CLINC-Full and CLINC-Imbal datasets (p <0.01). \textbf{Bayes.} represents our proposed Bayesian approximation via Monte-Carlo Dropout.}
\label{main_results}
\vspace{-0.4cm}
\end{table*}

% \vspace{-0.37cm}
\section{Experiments}
% \textbf{Datasets} We use two benchmark OOD datasets, CLINC-Full and CLINC-Imbal \cite{larson-etal-2019-evaluation}. \textbf{Metrics} We report IND metrics: Accuracy(Acc) and F1, and OOD metrics: Recall and F1. OOD Recall and F1 are the main evaluation metrics in this paper. \textbf{Baselines} We use MSP, LOF, GDA and Entropy for OOD detection, none of them need OOD supervised training. For feature extractor, we use LSTM and BERT. We present dataset statistics, baselines and details in the appendix. We will release our code after blind review. %强调比较的检测方法都是不需要OOD监督数据的
\subsection{Datasets}
\label{datasetdetails}
\begin{table}[h]
\centering
\resizebox{0.42\textwidth}{!}{%
\begin{tabular}{l|cc}
\hline
CLINC               & Full       & Imbal \\ \hline
Avg utterance length & 9          & 9           \\
Intents              & 150        & 150         \\
Training set size    & 15100      & 10625        \\
Training samples per class  & 100      & 25/50/75/100 \\
Training OOD samples amount & 100      & 100 \\
Development set size & 3100       & 3100        \\
Development samples per class & 20       & 20        \\
Development OOD samples amount & 100       & 100 \\
Testing Set Size     & 5500       & 5500        \\
Testing samples per class & 30       & 30        \\
Development OOD samples amount & 1000       & 1000 \\ \hline
\vspace{-0.5cm}
\end{tabular}
}
\vspace{-0.1cm}
\caption{Statistics of the CLINC datasets.}
\label{dataset}
% \vspace{-0.3cm}
\end{table}
We perform experiments on two public benchmark OOD datasets\footnote{https://github.com/clinc/oos-eval}, CLINC-Full and CLINC-Imbal \cite{larson-etal-2019-evaluation}. We show the detailed statistic of these datasets in Table \ref{dataset}. They both contain 150 in-domain intents across 10 domains. The only difference is that, for CLINC-Imbal, there are either 25, 50, 75 or 100 training queries per in-scope intent, rather than 100. Note that all the datasets we used have a fixed set of labeled OOD data but we don't use it for training.

\subsection{Metrics}
% \vspace{-0.1cm}
We report both OOD metrics: Recall and F1-score(F1) and in-domain metrics: F1-score(F1) and Accuracy(ACC). Since we aim to improve the performance of detecting out-of-domain intents from user queries, OOD Recall and F1 are the main evaluation metrics in this paper.

\subsection{Baselines}
% \vspace{-0.1cm}
% We use MSP, LOF, GDA and Entropy for OOD detection, none of them need OOD supervised training. For feature extractor, we use LSTM and BERT. We provide a more comprehensive comparison and implementation details of these models in the Appendix.

For detection algorithms, we use LOF, GDA, MSP and Entropy, none of them need OOD supervised training. For the feature extractor, we use LSTM and BERT. We provide a more comprehensive comparison and implementation details of these models in the Appendix.

\subsection{Main Results}
% \vspace{-0.1cm}
%0. Entropy在无监督检测方法中取得了最佳的性能，证明IND和OOD数据在类别分布的不确定性上存在较大的区别，
%1. 在基于概率的检测方法中，MSP大幅劣于Entropy，证明原始的OOD分布无法支持MSP的假设
%2. MC dropout对于softmax-based的检测方法能够带来一致性的提升，在MSP上更加显著，说明我们的方法有效缓解了OOD数据的over-confidence问题
%3. 此外，BERT特征提取器相较LSTM有微小的提升，IND训练集的不平衡会对IND分类和OOD检测的性能造成不良影响
% Table \ref{main_results} shows our main results. Among all the unsupervised OOD detection methods, \textbf{Entropy} obtained the best performance, which shows that there is a big difference in the uncertainty of the classes distribution between IND and OOD data. In the softmax-based detection methods, \textbf{MSP} is significantly inferior to \textbf{Entropy}, which proves that the original OOD distribution cannot support the assumption of \textbf{MSP}. After adding the MC dropout process, the softmax-based detection methods show consistently improvement, and the improvement in \textbf{MSP} is more significant, which shows that our method effectively alleviates the over-confidence problem of OOD data. In addition, the BERT feature extractor is slightly improved compared to LSTM, and the imbalance of the IND training set will adversely affect the performance of IND classification and OOD detection.

Table \ref{main_results} shows our main results on two benchmarks. Our Bayesian method significantly outperforms softmax-based baselines including MSP and Entropy, even distance-based SOTA GDA on OOD metrics. Specifically, on CLINC-Full, Bayes improves 19.92\% and 3.97\% OOD F1 compared to MSP and Entropy using LSTM, which proves MSP suffers from severe overconfidence and our method helps calibrate OOD distribution. The performance gap between MSP and Entropy is because Entropy based on softmax output distribution can better capture distinguished information for OOD than MSP based on a single value of softmax distribution. We find similar improvements under the BERT setting on CLINIC-Imbal dataset.
\begin{figure}[t]
    \centering
    \resizebox{.45\textwidth}{!}{
    \includegraphics{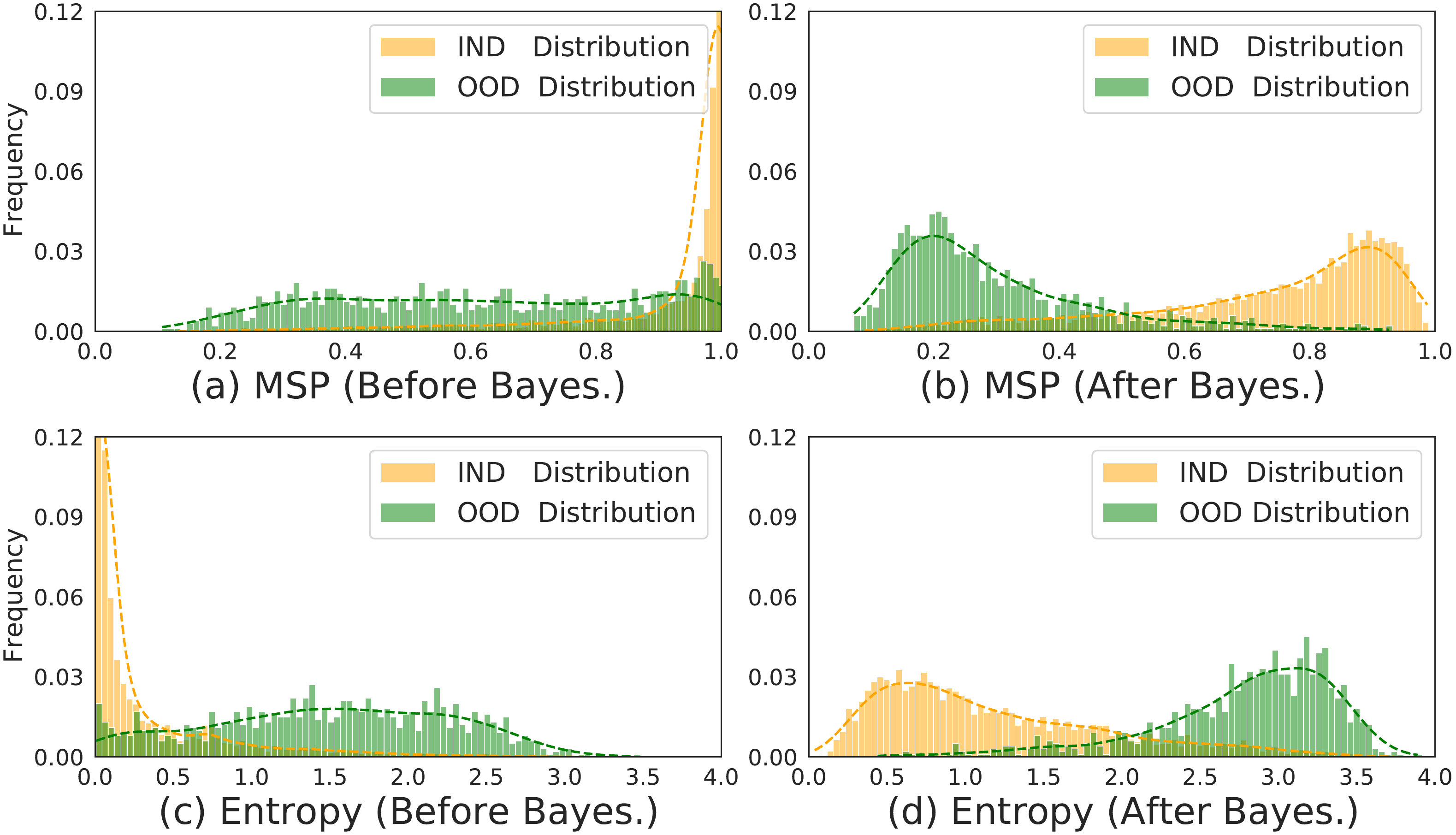}}
    \vspace{-0.2cm}
    \caption{Effect of Bayesian approximation on MSP and Entropy confidence distributions of IND and OOD.}
    \label{ana_dist}
     \vspace{-0.3cm}
\end{figure}

\section{Analysis}
%证明MC dropout如何提升OOD检测性能
\subsection{Effect of Bayesian approximation}
% \noindent\textbf{Effect of Bayesian approximation.} 
Fig \ref{ana_dist} shows the MSP and Entropy confidence distributions of IND and OOD test data using Bayesian to verify the effect of our method. Due to the over-confidence issue of OOD, we find IND and OOD curves overlap a lot in the original confidence scores. The overlapping part of Entropy is less, which confirms its better OOD detection performance. After calibration using Bayes, the overlap part of both methods is reduced, making it easier to distinguish between IND and OOD. 

% In this section, we explain why Bayesian approximation optimizes the softmax-based detection methods. We show in Figure \ref{ana_dist} the MSP and Entropy confidence score changes on IND data and OOD data, before and after Bayes. The data shown in the figure use the same feature extractor, obtained on the test set under a fixed dropout probability and sampling times. Due to the over-confidence phenomenon of OOD, we find IND and OOD overlap a lot in the original confidence score. The overlapping part of \textbf{Entropy} is less, which confirms its better OOD detection performance. After calibration, the overlap part of both methods is reduced, making it easier to distinguish between IND and OOD.

%分析MC dropout是否实现了OOD数据向均匀分布的校准
\subsection{Analysis of Distribution Calibration} 
Table \ref{ana_kl} shows the effect of Bayes on OOD Dirichlet Distribution. We calculate the KL-divergence between the predicted averaged softmax distribution and the uniform distribution of each test OOD sample and report the mean and median values on the whole test set. With the increase of sampling, we observe a larger drop on OOD mean and median KL values than INDs. It proves that Bayes can gradually calibrate the $v_{ood}$ to a uniform distribution and thus make the sparse OOD Dirichlet distribution dense but not affect IND. Besides, we find N = 100 already achieves good performance to reduce inference cost. We provide an efficiency comparison in Section \ref{cost} and find
33.33\% OOD F1 improvements only increase 0.41\% time.

% IND的分布不确定性不严重
%We find \textbf{MSP} for both IND and OOD data concentrate on high values, resulting in severe overconfidence. By contrast, energy scores better distinguish score distribution of OOD data from IND data. And energy distributions are smoother than softmax score distributions. Overall our proposed energy-based score function can disentangle confidence score distributions for IND and OOD data.

\subsection{Analysis of Cost-effectiveness}
% \vspace{-0.1cm}
\label{cost}
% 性能与资源的权衡？如何体现高性价比？
%     - 训练集包含15100条数据，测试集包含5500条数据。
%     - 传统的方法(整个训练+测试过程）所需的时间为4分钟，性能为50.13 ；
%     - 我们的方法dropout 10次时，所需要的额外时间为1.00s（+0.4%），性能为66.84（+33.3%）;【高性价比】
%     - 当dropout 100次，所需要的额外时间为12s，性能为70.05（+39.7%）;【高性价比】
%     - 当dropout 1000次时，所需要的额外时间是 152s，性能为70.82（+41.3%）；
% 当空间允许情况下，我们可以concat 多个sentence x；在时间允许的情况下，我们可以dropout n+次；
We show the comparison between the time consumption and the corresponding performance improvement in Table \ref{time} on CLINC-Full which has 15100 training data and 5500 test data. We find that when the number of samples N is 10, our method can improve the performance by 33.33\% while only increasing the time by 0.41\%, which proves that our proposed method is very cost-effective. Besides, we also find that more sampling times lead to more improvements, demonstrating that more accurate calibration significantly boosts OOD detection. In terms of time consumption and performance improvement, N = 100 is the most appropriate sampling parameter. When the sampling time is 1000, the cost-effectiveness is not as high as when $N = 10$. We consider that methods such as model distillation and pruning can reduce the time consumption, and we will leave it to future work. In general, we can choose the appropriate number of samples according to the computing resources.

\begin{table}[t]
\small
\centering
\resizebox{0.72\linewidth}{!}{
\begin{tabular}{c|c|c|c|c}
\toprule[1pt]
\multirow{3}{*}{$\lg$ $N$} & \multicolumn{4}{c}{Statistical Indicators} \\ \cline{2-5} 
\rule{0pt}{9pt} &\multicolumn{2}{c|}{OOD}&\multicolumn{2}{c}{IND} \\ \cline{2-3}\cline{4-5}
        ~ & mean & median & mean & median \\ \hline
        0 & 3.63  & 3.60  & 4.74  & 4.95  \\ \hline
        1 & 2.59  & 2.47  & 4.36  & 4.58  \\ \hline
        2 & 2.27  & 2.15  & 4.31  & 4.54  \\ \hline
        3 & 2.24  & 2.12 & 4.30  & 4.54 \\ \bottomrule[1pt]
\end{tabular}}
\vspace{-0.3cm}
\caption{KL-divergence between predicted distribution and uniform distribution on CLINC-Full. The smaller value is better for OOD. $N$ is the number of dropout.}
\label{ana_kl}
\vspace{-0.2cm}
\end{table}

\begin{table}[t]
\centering
\resizebox{0.87\linewidth}{!}{
    \begin{tabular}{c|c|c|c|c}
    \hline
        \multirow{2}{*}{lgN} & \multirow{2}{*}{Time(s)} & \multirow{2}{*}{OOD F1} & \multicolumn{2}{c}{Increased} \\ \cline{4-5}
        & & & Time(\%) & OOD F1(\%) \\ \hline
        0 & 240.00  & 50.13  & - & - \\ \hline
        1 & 240.98  & 66.84  & $\uparrow$ 0.41 & $\uparrow$ 33.33 \\ \hline
        2 & 252.41  & 70.05  & $\uparrow$ 5.17 & $\uparrow$ 39.74 \\ \hline
        3 & 388.36  & 70.82  & $\uparrow$ 61.82 & $\uparrow$ 41.27 \\ \hline
    \end{tabular}}
    \vspace{-0.2cm}
    \caption{Time consumption and corresponding performance improvement of Bayesian approximation based MSP.}
    % \vspace{-0.4cm}
\label{time}
\end{table}

\subsection{Analysis of Parameters} 
%分析MC dropout内部参数对于OOD检测的性能影响趋势
% dropout p和mcrourd比例，加drupout降低mcround次数。 +0.8结果。
Table \ref{ana_mc_param} reports the OOD F1 under different dropout probability and sampling times. Within a range between 0.3 to 0.7, the larger dropout probability leads to better OOD detection performance. This is because OOD data is more vulnerable to feature loss and its averaged softmax prediction distribution tends to be more uniform. Besides, more sampling times lead to improvements, demonstrating that more accurate calibration significantly boosts OOD detection. We also find that the performance on p=0.7, N=10 is better than the performance on p=0.3, N=1000. This prompts us to choose a higher p (e.g. 0.7), which can effectively reduce the time consumption (1000->10). In addition, OOD F1 is not sensitive to excessive sampling times.
\begin{table}[t]
\centering
\resizebox{0.89\linewidth}{!}{
\begin{tabular}{c|c|c|c|c|c|c}
\toprule[1pt]
\multirow{2}{*}{$\lg$ $N$} & \multicolumn{6}{c}{Dropout Probability} \\ \cline{2-7} 
                     & 0.3    & 0.4    & 0.5    & 0.6   & 0.7  & 0.8 \\ \hline
1                    & 57.21  & 60.97  & 61.80  & 64.27 & 66.84 & 64.01 \\ \hline
2                    & 60.78  & 64.38  & 65.84  & 68.87 & 70.05 & 69.03 \\ \hline
3                    & 63.32  & 65.35  & 67.51  & 69.33 & 70.82 & 69.69 \\ \bottomrule[1pt]
\end{tabular}}
\vspace{-0.2cm}
\caption{Effect of Bayesian approximation with different parameters on OOD F1-score.}
% \vspace{-0.5cm}
\label{ana_mc_param}
\end{table}

\vspace{0.2cm}
\section{Conclusion}
% \vspace{0.1cm}
In this paper, we conduct an analysis of why previous softmax-based detection algorithms like MSP or Entropy suffer from the overconfidence issue. We find OOD samples exactly yield a sparse distribution over the simplex and evenly distribute over the whole space.
% We find these methods have a wrong assumption that an OOD sample should have a flat softmax prediction distribution as close to the uniform distribution as possible. However, OOD samples exactly yield a sparse distribution over the simplex and evenly distributes over the whole space.
Therefore, we propose a simple but strong Bayesian approximation method to calibrate OOD distribution. Experiments prove the effectiveness of our method. We hope to provide new guidance for future OOD detection work.

\vspace{0.2cm}
\section*{Acknowledgements}
We thank all anonymous reviewers for their helpful comments and suggestions. This work was partially supported by MoE-CMCC "Artifical Intelligence" Project No. MCM20190701, National Key R\&D Program of China No. 2019YFF0303300 and Subject II No. 2019YFF0303302, DOCOMO Beijing Communications Laboratories Co., Ltd.

\bibliography{anthology,custom}

\begin{thebibliography}{23}
\expandafter\ifx\csname natexlab\endcsname\relax\def\natexlab#1{#1}\fi

\bibitem[{Akasaki and Kaji(2017)}]{Akasaki2017ChatDI}
Satoshi Akasaki and Nobuhiro Kaji. 2017.
\newblock Chat detection in an intelligent assistant: Combining task-oriented
  and non-task-oriented spoken dialogue systems.
\newblock \emph{ArXiv}, abs/1705.00746.

\bibitem[{Devlin et~al.(2019)Devlin, Chang, Lee, and
  Toutanova}]{Devlin2019BERTPO}
Jacob Devlin, Ming-Wei Chang, Kenton Lee, and Kristina Toutanova. 2019.
\newblock \href {https://doi.org/10.18653/v1/N19-1423} {{BERT}: Pre-training of
  deep bidirectional transformers for language understanding}.
\newblock In \emph{Proceedings of the 2019 Conference of the North {A}merican
  Chapter of the Association for Computational Linguistics: Human Language
  Technologies, Volume 1 (Long and Short Papers)}, pages 4171--4186,
  Minneapolis, Minnesota. Association for Computational Linguistics.

\bibitem[{Gal and Ghahramani(2016)}]{pmlr-v48-gal16}
Yarin Gal and Zoubin Ghahramani. 2016.
\newblock \href {http://proceedings.mlr.press/v48/gal16.html} {Dropout as a
  bayesian approximation: Representing model uncertainty in deep learning}.
\newblock In \emph{Proceedings of The 33rd International Conference on Machine
  Learning}, volume~48 of \emph{Proceedings of Machine Learning Research},
  pages 1050--1059, New York, New York, USA. PMLR.

\bibitem[{Gnewuch et~al.(2017)Gnewuch, Morana, and
  Maedche}]{Gnewuch2017TowardsDC}
Ulrich Gnewuch, S.~Morana, and A.~Maedche. 2017.
\newblock Towards designing cooperative and social conversational agents for
  customer service.
\newblock In \emph{International Conference on Information Systems}.

\bibitem[{Guo et~al.(2017)Guo, Pleiss, Sun, and Weinberger}]{Guo2017OnCO}
Chuan Guo, Geoff Pleiss, Yu~Sun, and Kilian~Q Weinberger. 2017.
\newblock On calibration of modern neural networks.
\newblock In \emph{International Conference on Machine Learning}, pages
  1321--1330. PMLR.

\bibitem[{Hendrycks and Gimpel(2017)}]{Hendrycks2017ABF}
Dan Hendrycks and Kevin Gimpel. 2017.
\newblock A baseline for detecting misclassified and out-of-distribution
  examples in neural networks.
\newblock \emph{ArXiv}, abs/1610.02136.

\bibitem[{Hochreiter and Schmidhuber(1997)}]{Hochreiter1997LongSM}
S.~Hochreiter and J.~Schmidhuber. 1997.
\newblock Long short-term memory.
\newblock \emph{Neural Computation}, 9:1735--1780.

\bibitem[{Kingma and Ba(2014)}]{kingma2014adam}
Diederik~P Kingma and Jimmy Ba. 2014.
\newblock Adam: A method for stochastic optimization.
\newblock \emph{arXiv preprint arXiv:1412.6980}.

\bibitem[{Kiureghian and Ditlevsen(2009)}]{twouncertainty2009}
Armen~Der Kiureghian and O.~Ditlevsen. 2009.
\newblock Aleatory or epistemic? does it matter?
\newblock \emph{Structural Safety}, 31:105--112.

\bibitem[{Larson et~al.(2019)Larson, Mahendran, Peper, Clarke, Lee, Hill,
  Kummerfeld, Leach, Laurenzano, Tang, and Mars}]{larson-etal-2019-evaluation}
Stefan Larson, Anish Mahendran, Joseph~J. Peper, Christopher Clarke, Andrew
  Lee, Parker Hill, Jonathan~K. Kummerfeld, Kevin Leach, Michael~A. Laurenzano,
  Lingjia Tang, and Jason Mars. 2019.
\newblock \href {https://www.aclweb.org/anthology/D19-1131} {An evaluation
  dataset for intent classification and out-of-scope prediction}.
\newblock In \emph{Proceedings of the 2019 Conference on Empirical Methods in
  Natural Language Processing and the 9th International Joint Conference on
  Natural Language Processing (EMNLP-IJCNLP)}.

\bibitem[{Liang et~al.(2018)Liang, Li, and Srikant}]{Liang2018EnhancingTR}
Shiyu Liang, Yixuan Li, and R.~Srikant. 2018.
\newblock Enhancing the reliability of out-of-distribution image detection in
  neural networks.
\newblock \emph{arXiv: Learning}.

\bibitem[{Lin and Xu(2019)}]{Lin2019DeepUI}
Ting-En Lin and Hua Xu. 2019.
\newblock Deep unknown intent detection with margin loss.
\newblock In \emph{Proceedings of the 57th Annual Meeting of the Association
  for Computational Linguistics}, pages 5491--5496.

\bibitem[{Malinin and Gales(2018)}]{Malinin2018PredictiveUE}
A.~Malinin and M.~Gales. 2018.
\newblock Predictive uncertainty estimation via prior networks.
\newblock In \emph{Proceedings of the 32nd International Conference on Neural
  Information Processing Systems}, page 7047–7058.

\bibitem[{Pennington et~al.(2014)Pennington, Socher, and
  Manning}]{pennington2014glove}
Jeffrey Pennington, Richard Socher, and Christopher~D Manning. 2014.
\newblock Glove: Global vectors for word representation.
\newblock In \emph{Proceedings of the 2014 conference on empirical methods in
  natural language processing (EMNLP)}, pages 1532--1543.

\bibitem[{Shum et~al.(2018)Shum, He, and Li}]{Shum2018FromET}
H.~Shum, X.~He, and Di~Li. 2018.
\newblock From eliza to xiaoice: challenges and opportunities with social
  chatbots.
\newblock \emph{Frontiers of Information Technology \& Electronic Engineering},
  19:10--26.

\bibitem[{Tsymbalov et~al.(2020)Tsymbalov, Fedyanin, and
  Panov}]{Tsymbalov2020DropoutSB}
Evgenii Tsymbalov, K.~Fedyanin, and Maxim Panov. 2020.
\newblock Dropout strikes back: Improved uncertainty estimation via diversity
  sampled implicit ensembles.
\newblock \emph{ArXiv}, abs/2003.03274.

\bibitem[{Tulshan and Dhage(2018)}]{Tulshan2018SurveyOV}
Amrita~S Tulshan and Sudhir~Namdeorao Dhage. 2018.
\newblock Survey on virtual assistant: Google assistant, siri, cortana, alexa.
\newblock In \emph{International symposium on signal processing and intelligent
  recognition systems}, pages 190--201.

\bibitem[{Wu et~al.(2022)Wu, He, Yan, Gao, Zeng, Zheng, Zhao, Jiang, Wu, and
  Xu}]{wu-etal-2022-revisit}
Yanan Wu, Keqing He, Yuanmeng Yan, QiXiang Gao, Zhiyuan Zeng, Fujia Zheng, Lulu
  Zhao, Huixing Jiang, Wei Wu, and Weiran Xu. 2022.
\newblock \href {https://doi.org/10.18653/v1/2022.naacl-main.307} {Revisit
  overconfidence for {OOD} detection: Reassigned contrastive learning with
  adaptive class-dependent threshold}.
\newblock In \emph{Proceedings of the 2022 Conference of the North American
  Chapter of the Association for Computational Linguistics: Human Language
  Technologies}, pages 4165--4179, Seattle, United States. Association for
  Computational Linguistics.

\bibitem[{Xu et~al.(2020)Xu, He, Yan, Liu, Liu, and Xu}]{xu-etal-2020-deep}
Hong Xu, Keqing He, Yuanmeng Yan, Sihong Liu, Zijun Liu, and Weiran Xu. 2020.
\newblock \href {https://doi.org/10.18653/v1/2020.coling-main.125} {A deep
  generative distance-based classifier for out-of-domain detection with
  mahalanobis space}.
\newblock In \emph{Proceedings of the 28th International Conference on
  Computational Linguistics}, pages 1452--1460, Barcelona, Spain (Online).
  International Committee on Computational Linguistics.

\bibitem[{Yao and Gao(2016)}]{Yao2016LawOL}
Kai Yao and Jinwu Gao. 2016.
\newblock Law of large numbers for uncertain random variables.
\newblock \emph{IEEE Transactions on Fuzzy Systems}, 24:615--621.

\bibitem[{Zeng et~al.(2021{\natexlab{a}})Zeng, He, Yan, Liu, Wu, Xu, Jiang, and
  Xu}]{zeng-etal-2021-modeling}
Zhiyuan Zeng, Keqing He, Yuanmeng Yan, Zijun Liu, Yanan Wu, Hong Xu, Huixing
  Jiang, and Weiran Xu. 2021{\natexlab{a}}.
\newblock \href {https://doi.org/10.18653/v1/2021.acl-short.110} {Modeling
  discriminative representations for out-of-domain detection with supervised
  contrastive learning}.
\newblock In \emph{Proceedings of the 59th Annual Meeting of the Association
  for Computational Linguistics and the 11th International Joint Conference on
  Natural Language Processing (Volume 2: Short Papers)}, pages 870--878,
  Online. Association for Computational Linguistics.

\bibitem[{Zeng et~al.(2021{\natexlab{b}})Zeng, He, Yan, Xu, and
  Xu}]{zeng-etal-2021-adversarial}
Zhiyuan Zeng, Keqing He, Yuanmeng Yan, Hong Xu, and Weiran Xu.
  2021{\natexlab{b}}.
\newblock \href {https://doi.org/10.18653/v1/2021.naacl-main.447} {Adversarial
  self-supervised learning for out-of-domain detection}.
\newblock In \emph{Proceedings of the 2021 Conference of the North American
  Chapter of the Association for Computational Linguistics: Human Language
  Technologies}, pages 5631--5639, Online. Association for Computational
  Linguistics.

\bibitem[{Zheng et~al.(2020)Zheng, Chen, and Huang}]{Zheng2020OutofDomainDF}
Yinhe Zheng, Guanyi Chen, and Minlie Huang. 2020.
\newblock Out-of-domain detection for natural language understanding in dialog
  systems.
\newblock \emph{IEEE/ACM Transactions on Audio, Speech, and Language
  Processing}, 28:1198--1209.

\end{thebibliography}
\bibliographystyle{acl_natbib}

\appendix

% \section{Dataset Details}
% \label{datasetdetails}

% Table \ref{dataset} shows the details of two benchmark OOD dataset\footnote{https://github.com/clinc/oos-eval} CLINC-Full and CLINC-Imbal \cite{Larson2019AnED}. They both contain 150 in-domain intents across 10 domains. The only difference is that, for CLINC-Imbal, there are either 25, 50, 75 or 100 training queries per in-scope intent, rather than 100.

% \begin{table}[h]
% \centering
% \resizebox{0.48\textwidth}{!}{%
% \begin{tabular}{l|cc}
% \hline
% CLINC               & Full       & Imbal \\ \hline
% Avg utterance length & 9          & 9           \\
% Intents              & 150        & 150         \\
% Training set size    & 15100      & 10625        \\
% Training samples per class  & 100      & 25/50/75/100 \\
% Training OOD samples amount & 100      & 100 \\
% Development set size & 3100       & 3100        \\
% Development samples per class & 20       & 20        \\
% Development OOD samples amount & 100       & 100 \\
% Testing Set Size     & 5500       & 5500        \\
% Testing samples per class & 30       & 30        \\
% Development OOD samples amount & 1000       & 1000 \\ \hline
% \end{tabular}
% }
% \caption{Statistics of the CLINC datasets.}
% \label{dataset}
% \end{table}

\section{Baseline Details}
% 强调我们是无监督，采取了 MSP GDA LOF Entropy 四种检测方法，详细介绍GDA和LOF即可，再简要介绍一下BiLSTM和BERT。
We compare many types of unsupervised OOD detection models. For detection algorithms, we use LOF(Local Outlier Factor)\cite{Lin2019DeepUI}, GDA(Gaussian Discriminant Analysis)\cite{xu-etal-2020-deep}, MSP(Maximum Softmax Probability)\cite{Hendrycks2017ABF} and Entropy. For feature extractor, we use LSTM(Long Short Term Memory)\cite{Hochreiter1997LongSM} and BERT(Bidirectional Encoder Representations from Transformers)\cite{Devlin2019BERTPO}. 

\noindent\textbf{MSP} (Maximum Softmax Probability)\cite{Hendrycks2017ABF} uses maximum softmax probability as the confidence score and regards an intent as OOD if the score is below a fixed threshold.

\textbf{LOF} (Local Outlier Factor)\cite{Lin2019DeepUI} A detecting unknown intents in the utterance algorithm with local density. It Assumes that unknown intents' local density is significantly lower than its k-nearest neighbor's.

\textbf{GDA} (Gaussian Discriminant Analysis) \cite{xu-etal-2020-deep} A generative distance-based classifier for OOD detection with Euclidian space. For avoiding over-confidence problems, they estimate the class-conditional distribution on feature spaces of DNNs via Gaussian discriminant analysis. GDA is the state-of-the-art detection method till now, our proposed method using Bayesian approximation still significantly outperforms GDA. We also compare our method on two feature extractors for further study. 

\textbf{LSTM} (Long Short Term Memory)\cite{Hochreiter1997LongSM} A neural network that was proposed with the motivation of an analysis of Recurrent Neural Nets, which found that long time lags were inaccessible to existing architectures because backpropagated error either blows up or decays exponentially. 

\textbf{BERT} (Bidirectional Encoder Representations from Transformers)\cite{Devlin2019BERTPO} A neural network that is trained to predict elided words in the text and then fine-tuned on our data. Note that they both trained only on labeled in-domain data using cross-entropy loss.

\begin{figure}[t]
    \centering
    \resizebox{.48\textwidth}{!}{
    \includegraphics{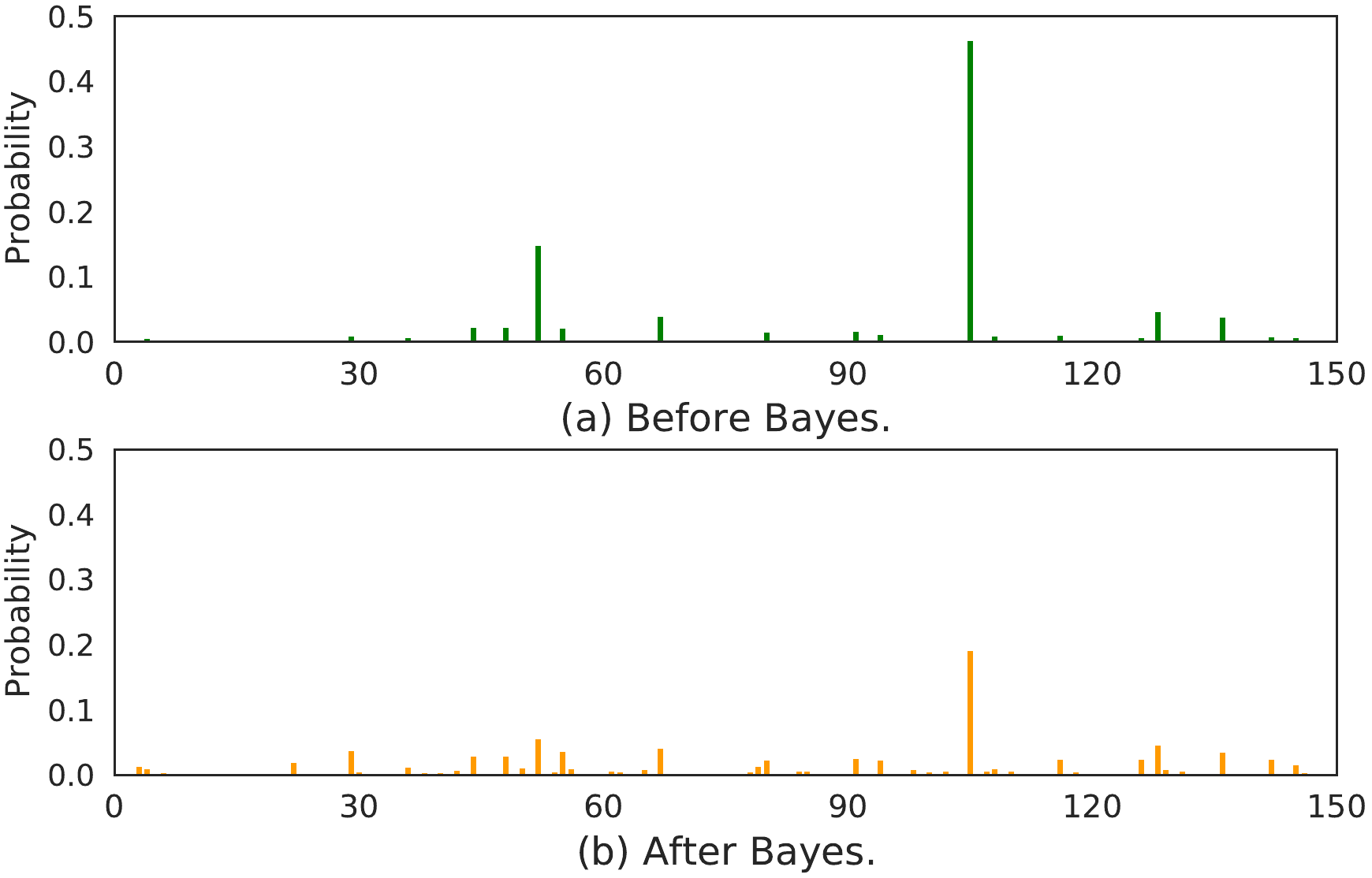}}
    \vspace{-0.7cm}
    \caption{Effect of Bayesian approximation on softmax distribution of OOD sample.}
    \label{ana_vis}
     \vspace{-0.5cm}
\end{figure}

\section{Implementation Details}
We use the public pre-trained 300 dimensions GloVe embeddings \cite{pennington2014glove}\footnote{https://github.com/stanfordnlp/GloVe} or bert-base-uncased \cite{Devlin2019BERTPO}\footnote{https://github.com/google-research/bert} model to embed tokens. We use a two-layer BiLSTM as a feature extractor and set the dimension of hidden states to 128. We use Adam optimizer \cite{kingma2014adam} to train our model. We set a learning rate to 1E-03 for GloVe+LSTM and 1E-04 for BERT. In the training stage, We set the dropout probability to 0.5 and set the training epoch up to 200 with an early stop. We train only on in-domain labeled data. We use the best F1 scores on the validation set to calculate the detection method's threshold adaptively. For our proposed Bayesian approximation, we set the dropout probability to 0.7, and the dropout sampling times to 100. Each result of the experiments is tested 10 times under the same setting and gets the average value. The training stage of our model lasts about 4 minutes using GloVe embeddings, and 12 minutes using Bert-base-uncased, both on a single Tesla T4 GPU(16 GB of memory). The average value of the trainable model parameters is 3.05M. We will release our code after blind review.

%直观展示MC dropout对样本分布的影响（辅助理解）
\section{Visualization of softmax prediction distribution.} 
\subsection{Visualization of OOD samples}
In Fig \ref{ana_vis}, we give a 150-dimensional class distribution of an OOD sample to help understand our calibration process. The upper part of the figure is the distribution obtained by using the primary feature extractor. The softmax prediction distribution has obvious over-confidence in a particular IND category. The lower half of the figure presents the distribution after calibration by Bayesian approximation, which is flatter and meets the expectations of the OOD sample. When applying softmax-based detection methods, the latter will more easily recognized as OOD.

\subsection{Visualization of IND samples}
\label{ind_random}
\begin{figure}[t]
    \centering
    \resizebox{.46\textwidth}{!}{
    \includegraphics{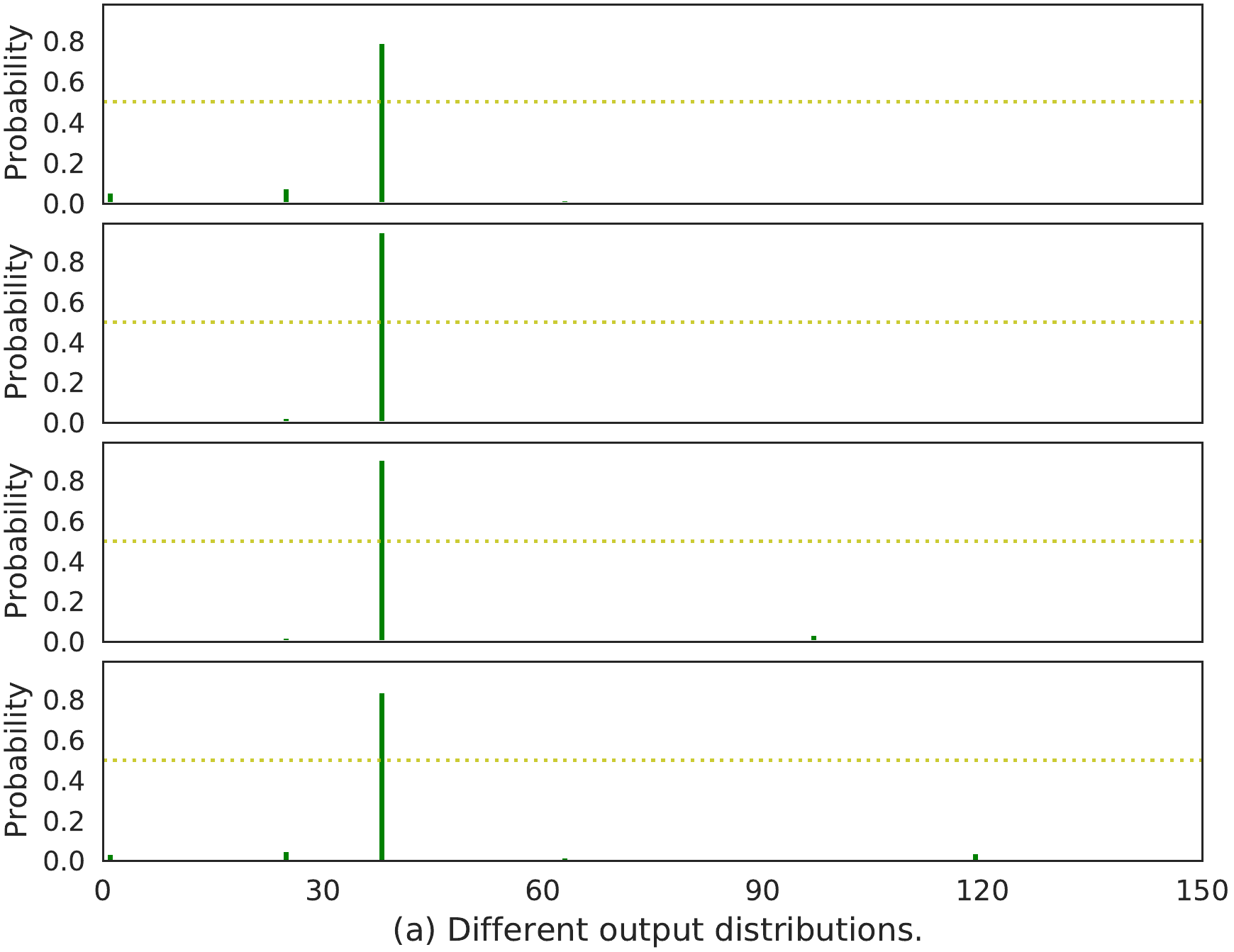}}
    \vspace{-0.3cm}
    \label{intro_seen_before}
     % \vspace{-0.1cm}
\end{figure}
\begin{figure}[t]
    \centering
    \resizebox{.46\textwidth}{!}{
    \includegraphics{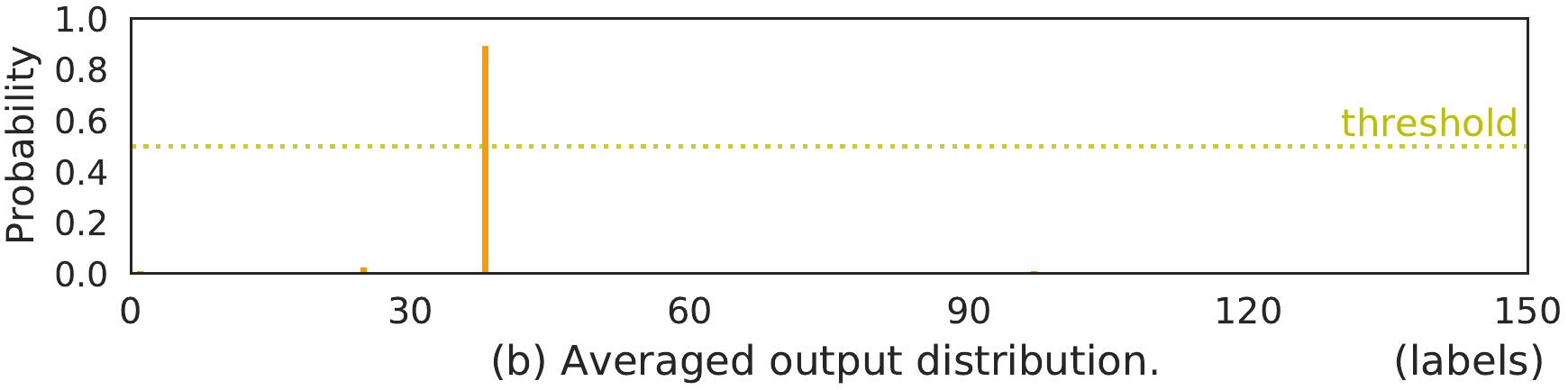}}
    % \vspace{-0.1cm}
    \caption{Multiple predicted probability distributions of an IND sample under different random seeds.}
    %Predicted probability distribution of an IND sample under different random seeds settings.
    \label{intro_seen_after}
\end{figure}

\begin{figure}[t]
    \centering
    \resizebox{.46\textwidth}{!}{
    \includegraphics{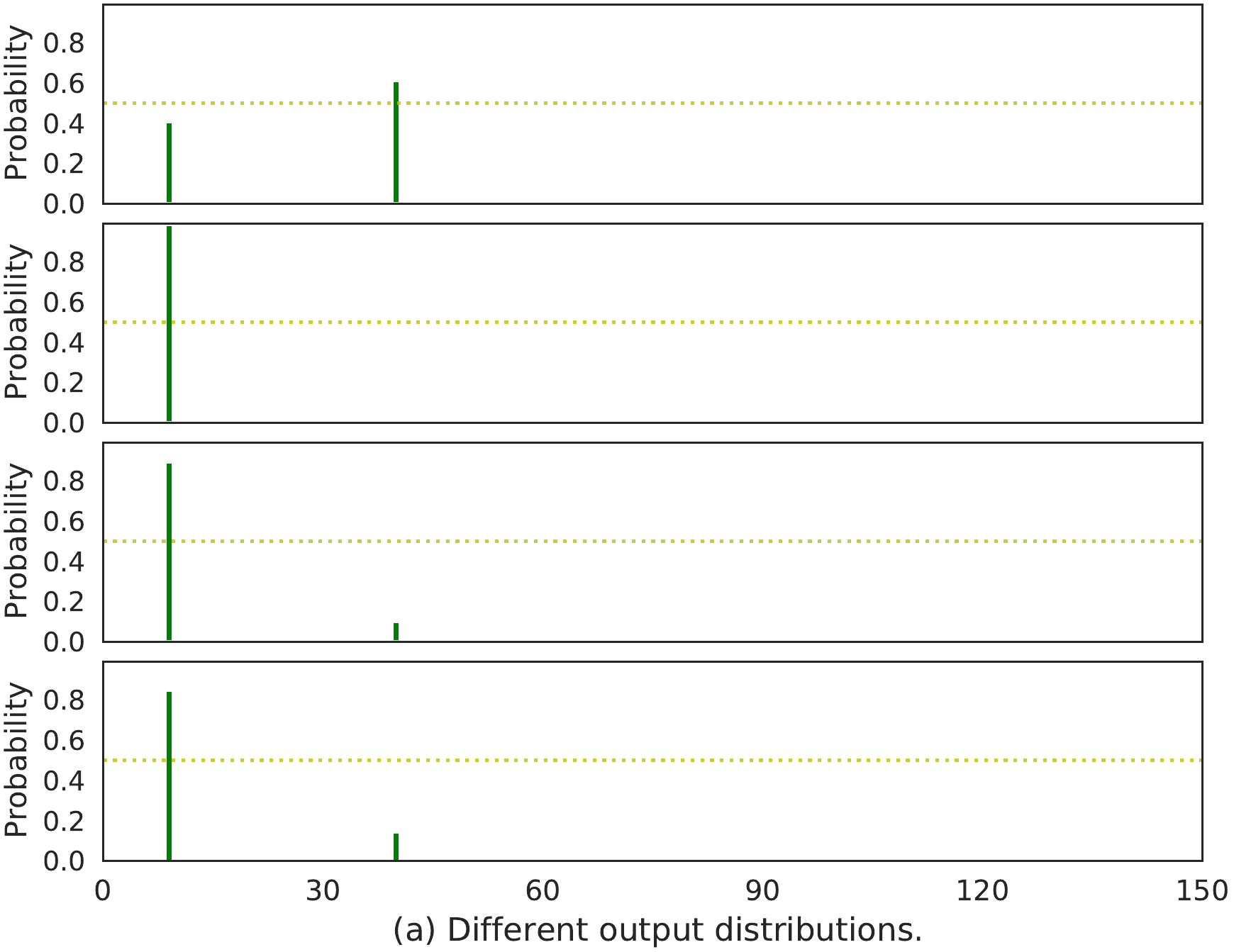}}
    \vspace{-0.25cm}
    \label{intro_seen_before_error}
     \vspace{-0.1cm}
\end{figure}
\begin{figure}[t]
    \centering
    \resizebox{.46\textwidth}{!}{
    \includegraphics{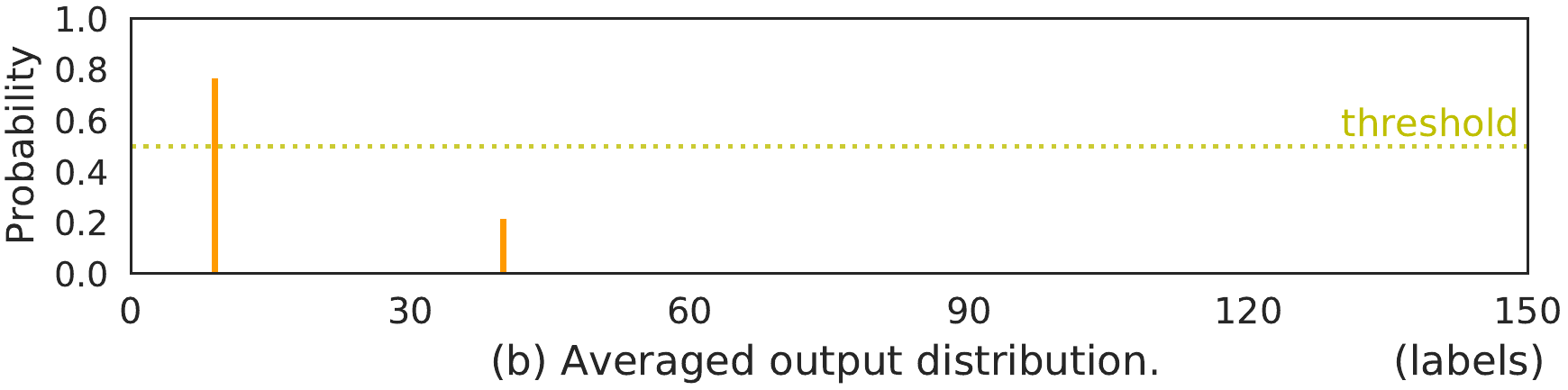}}
    % \vspace{-0.1cm}
    \caption{Multiple predicted probability distributions of an IND sample under different random seeds.}
    \label{intro_seen_after_error}
\end{figure}

% Multiple predicted probability distributions of an IND sample under different random seeds. We train four identical models on the same data but only use different random seeds. Fig (a) displays each output distribution of an IND input and Fig (b) shows the averaged output distribution.
Corresponding to Fig \ref{intro_unseen_after}, Fig \ref{intro_seen_after} and Fig \ref{intro_seen_after_error} show predicted probability distributions of two IND sample under different random seeds over 150 classes. We train four identical models on the same data but only use different random seeds. Fig (a) displays each output distribution of an IND input and Fig (b) shows the averaged output distribution.
Specifically, Fig \ref{intro_seen_after} shows when the input utterance is '\emph{block my american saving bank for now}', the model obtains the maximum prediction probability on ground truth (\emph{freeze\_account}) under four random seeds and averaged output. Specifically, The maximum probabilities of prediction are 0.77, 0.96, 0.90 and 0.84 under different random seeds, and 0.87 under averaged output. In the experiments, we find that most IND samples present the state of Fig \ref{intro_seen_after}, that is, under different random samples setting, the model is very confident to give the input IND utterances with high confidence probability in the ground-truth category. We guess that this is because the model has seen some IND data in the training phase, and is familiar with IND classification, that is, the distribution uncertainty of IND is not serious. 
We also show another example in Fig \ref{intro_seen_after_error} which the input utterance is '\emph{where is improve the credit score}' and the corresponding true label is \emph{improve\_credit\_score}. However, we find that under one random sampling setting, the model mispredicts into \emph{credit\_score} category with a probability of 0.61. We argue this is due to the fact that the two are easily confused with each other. Under this random sampling parameter setting, the model has not learned the feature ability to accurately distinguish these two categories. In addition, we also find that although there are wrong predictions, most of the IND predictions are accurate and have a high prediction probability, so that the highest prediction probability can still be obtained on the ground-truth label after averaging the distributions. This also reveals that our method will not damage the performance of IND classification, and can even avoid misjudgment among some confusing IND categories.

\end{document}